\documentclass[runningheads]{llncs}

\usepackage[year=2026]{eccv}
\usepackage{eccvabbrv}

\usepackage{graphicx}
\usepackage{amsmath}
\usepackage{amssymb}
\usepackage{booktabs}
\usepackage{multirow}
\usepackage{array}
\usepackage{makecell}
\usepackage{xcolor}
\usepackage{float}
\usepackage[breaklinks,colorlinks,citecolor=eccvblue]{hyperref}
\usepackage{orcidlink}

\graphicspath{{figures/}}

\newcommand{\resulttablesize}{\small}

\begin{document}
\raggedbottom

\title{Curating Same-Family Neural Networks for LLM-Guided Model Improvement: A Controlled Case Study}
\titlerunning{Curating Same-Family Neural Networks}

\author{Kabir Dev Paul Baghel\textsuperscript{*}\orcidlink{0009-0003-4428-4818} \and Radu Timofte \and Dmitry Ignatov}
\authorrunning{K. D. P. Baghel et al.}
\institute{Computer Vision Lab, CAIDAS, University of W\"urzburg, Germany\\
\email{kabirdpaulbaghel@gmail.com, \{radu.timofte, dmytro.ignatov\}@uni-wuerzburg.de}}
\maketitle
 
\begin{abstract}
\emergencystretch=2em
Neural-network repositories contain executable models, recipes, input transformations, and measured accuracies. We study whether one same-family experiment can be curated as prompt guidance for LLM-based improvement of a low-performing target under equal generation and evaluation budgets. TuneNNGen extends NNGPT with a source-guided route and compares it with target-only generation on one CIFAR-10 target, two fixed source-selection rules, three code LLMs, and an additional SVHN target. Under the historical one-epoch search protocol, best-of-budget accuracy on the available evaluation split rises from 23.98\% to 50.49\% on CIFAR-10 and from 22.54\% to 78.80\% on SVHN. Selected five-epoch, three-seed means on train-derived validation splits retain gains of 40.94 points on CIFAR-10, 18.83 on Imagenette, and 7.27 on CIFAR-100. Direct-copy and negative-control analyses show that gains depend on source-target compatibility and LLM adaptation; stored source accuracy alone does not predict transferability.
\end{abstract}
\section{Introduction}

Modern machine learning increasingly relies on large collections of previous experiments. Beyond datasets themselves, many repositories store executable neural-network implementations together with their training recipes, input transformations, measured accuracies, and other metadata. Such collections provide curated experimental evidence that can potentially guide future model development. When this information is available, improving a model under a fixed compute budget becomes not only an optimization problem but also an \emph{experiment-curation problem}: which previously evaluated experiment should be selected to guide the search for a better model?

Large language models (LLMs) provide a practical mechanism for exploiting curated experiment collections because a selected experiment can be placed directly into the prompt. Prior work in the ABrain/LEMUR ecosystem has explored LLMs for architecture generation, prompt engineering, hyperparameter optimization, and iterative neural architecture search with feedback memory~\cite{ABrain.NNGPT,ABrain.Prompt,ABrain.HPGPT,ABrain.Feedback_Memory}. Within this line of work, little attention has been paid to the selection step itself. A selected experiment contributes executable architecture code, training settings, input transformations, and measured performance, yet it remains unclear which stored experiment provides the most useful guidance for improving a different network.

This work studies deep model curation using the LEMUR Neural Network Database. Each database record represents a complete evaluated experiment containing executable model code, training settings, input transformations, and measured accuracy. Given a low-performing target network, we select one higher-performing experiment from the same task and architecture family and provide both records to the LLM. The selected experiment serves as curated evidence from which the LLM generates candidate improvements for the target.

Two questions motivate this study. First, does adding a curated same-family experiment improve equal-budget LLM-guided search compared with using the target network alone? Second, which characteristics make a stored experiment transferable? In particular, is the highest recorded accuracy the best selection criterion, or is a more closely related architecture a better source of guidance?

We answer these questions with equal-budget comparisons: target-only and source-guided generation receive identical generation and evaluation budgets. We compare target-only generation with a selected same-family source, compare two fixed source-selection rules for one CIFAR-10 target, and separate direct recipe transfer from additional LLM adaptation.

This paper addresses the following scientific question:

\begin{quote}
\emph{How should a single previously evaluated neural network be curated so that it most effectively guides equal-budget LLM-based improvement of a lower-performing target network?}
\end{quote}

We evaluate three complementary aspects of performance: generation reliability, measured by the number of successfully evaluated candidates; search effectiveness, measured by the best accuracy found within a fixed budget; and generation consistency, measured by the mean accuracy of successfully evaluated candidates.

\paragraph{Contributions.}
This paper makes the following contributions:

\begin{itemize}
    \item We formulate LLM-guided neural-network improvement as a \emph{neural-network curation} problem in which a single previously evaluated neural network is selected from a curated database to guide candidate generation.

    \item For one CIFAR-10 target, the architecture-related source gives a more favorable guided advantage than the highest-accuracy source for all three LLMs.

    \item We add a direct-copy condition splitting the gain into what the recorded recipe achieves unchanged and what the LLM adds: on CIFAR-10 \texttt{archbest}, direct copying contributes 18.69 points and LLM adaptation 7.82; on SVHN, the corresponding contributions are $-2.95$ and $+59.21$ points.

    \item We evaluate the approach across multiple datasets, a small set of architecture families, and code LLMs, and confirm the strongest recipe-guidance results through independent generations and selected validation-first multi-seed experiments.
\end{itemize}
\section{Related Work}

\paragraph{Automated architecture and recipe search.}
Neural architecture search (NAS) includes reinforcement learning~\cite{Zoph2016NAS}, evolution~\cite{Real2019RegularizedEvolution}, differentiable search~\cite{Liu2019DARTS}, and model scaling~\cite{Tan2019EfficientNet}; Elsken et al. provide a survey~\cite{Elsken2019SurveyNAS}. Hyperparameter optimization searches training recipes through methods such as sequential model-based optimization, Hyperband, and define-by-run systems~\cite{Bergstra2011Hyperparameter,Li2018Hyperband,Akiba2019Optuna}. These methods are particularly relevant because the strongest improvements in our experiments come from training settings. Warm-start and meta-learning methods also reuse observations from related tasks or prior studies to reduce the cost of a new search~\cite{Feurer2015AutoML}. Our setting is complementary: the transferred object is a complete executable neural-network record, and the LLM adapts it under an equal candidate budget. We follow NAS reproducibility guidance by fixing candidate budgets, validation rules, and target/source pairs~\cite{Lindauer2020BestPracticesNAS}.

\paragraph{LLM generation, retrieval, and optimization.}
Code LLMs can synthesize executable programs; sampling, filtering, and self-refinement can improve reliability~\cite{Chen2021Codex,Li2022AlphaCode,Madaan2023SelfRefine,Shinn2023Reflexion}. Few-shot and retrieval-augmented prompting use external examples or retrieved information to guide generation~\cite{Brown2020FewShot,Lewis2020RAG}. Which examples are chosen matters: prior work selects them by similarity or with trained retrievers~\cite{Liu2022GoodExamples,Rubin2022EPR}, whereas our examples are complete prior experiments, making architecture and recorded accuracy separable selection criteria. OPRO, EvoPrompt, and LLMatic place LLMs inside optimization loops~\cite{Yang2023OPRO,Guo2023EvoPrompt,Nasir2023LLMatic}. Our work applies this idea to neural-network improvement by selecting a higher-performing model from the same architecture family and using its recorded configuration as guidance.

\paragraph{ABrain/LEMUR context.}
The implementation builds on NNGPT, prompt-based architecture generation, LLM hyperparameter tuning, and feedback-memory search~\cite{ABrain.NNGPT,ABrain.Prompt,ABrain.HPGPT,ABrain.Feedback_Memory}. LEMUR supplies the network code, training recipes, input transformations, and measured accuracies used to form target/source pairs~\cite{ABrain.NN-Dataset,ABrain.LEMUR2}. TuneNNGen extends NNGPT with an equal-budget source-guided route: the LLM receives one curated LEMUR source record in addition to the target and proposes either a training recipe or a structured architecture edit; the target-only route omits the source. Exact prompts, configuration identifiers, and output schemas are provided in the Supplementary Material and artifact archive. This study compares source-guided generation with an equal-budget target-only search and then examines which prior experiment should be selected as prompt guidance. Fixed output formats define which parts of a candidate the LLM may change, making each generated candidate straightforward to validate and compare.

\section{Method}
\label{sec:method}

\subsection{Problem Setup}

We consider a setting where a database contains previously evaluated neural networks that can be used as guidance examples for LLM-based model improvement. Each database entry contains executable architecture code, training settings, input transformations, and measured accuracy. Given a low-accuracy target network, the goal is to select a single higher-performing source network whose information improves the quality of candidates generated under a fixed compute budget.

For a dataset $D$, let $T$ denote the target network and let $S$ denote a selected source network from the same task and architecture family, with $a(S)>a(T)$. The LLM generates candidate versions $T'_1,\ldots,T'_n$ of the target using either the target alone or the target together with the source example. The final selected model is

\begin{equation}
    T^* = \arg\max_{T'_i \in \mathcal{V}} a(T'_i),
\end{equation}

where $\mathcal{V}$ contains candidates that parse, satisfy the required software interface, train successfully, and return a measured accuracy.

The central methodological question is therefore not only how to generate candidates, but which previously evaluated network should be curated as the source example.

\subsection{Source Selection}

The source-selection stage determines which previously evaluated neural network is provided to the LLM as guidance. For each target, we select a higher-accuracy record from the same task and architecture family. The source contains the architecture implementation, training settings, input transformations, and recorded accuracy. The LLM uses this information as an example of a successful related model while generating candidates for the target.

We compare two curation strategies. The first selects the eligible record with the highest recorded accuracy. The second selects a high-performing record with a more closely related architecture. The comparison tests whether historical performance alone is sufficient for selecting useful guidance or whether structural similarity improves transfer.

In the reported CIFAR-10 experiment, both source records were fixed before candidate generation and were not selected by a learned retriever. \texttt{highsource} is the eligible record with the highest stored accuracy (81.94\%). It uses a residual bottleneck design. \texttt{archbest} has lower stored accuracy (74.27\%) but a closer architecture organization and was selected using the fixed manual criterion of matching the target's AlexNet-style layer organization: five convolutional and three linear layers. This is a controlled comparison of two fixed source records for one target, not evidence for a generally optimal retrieval rule.

\begin{figure}[!ht]
\centering
\includegraphics[width=\textwidth]{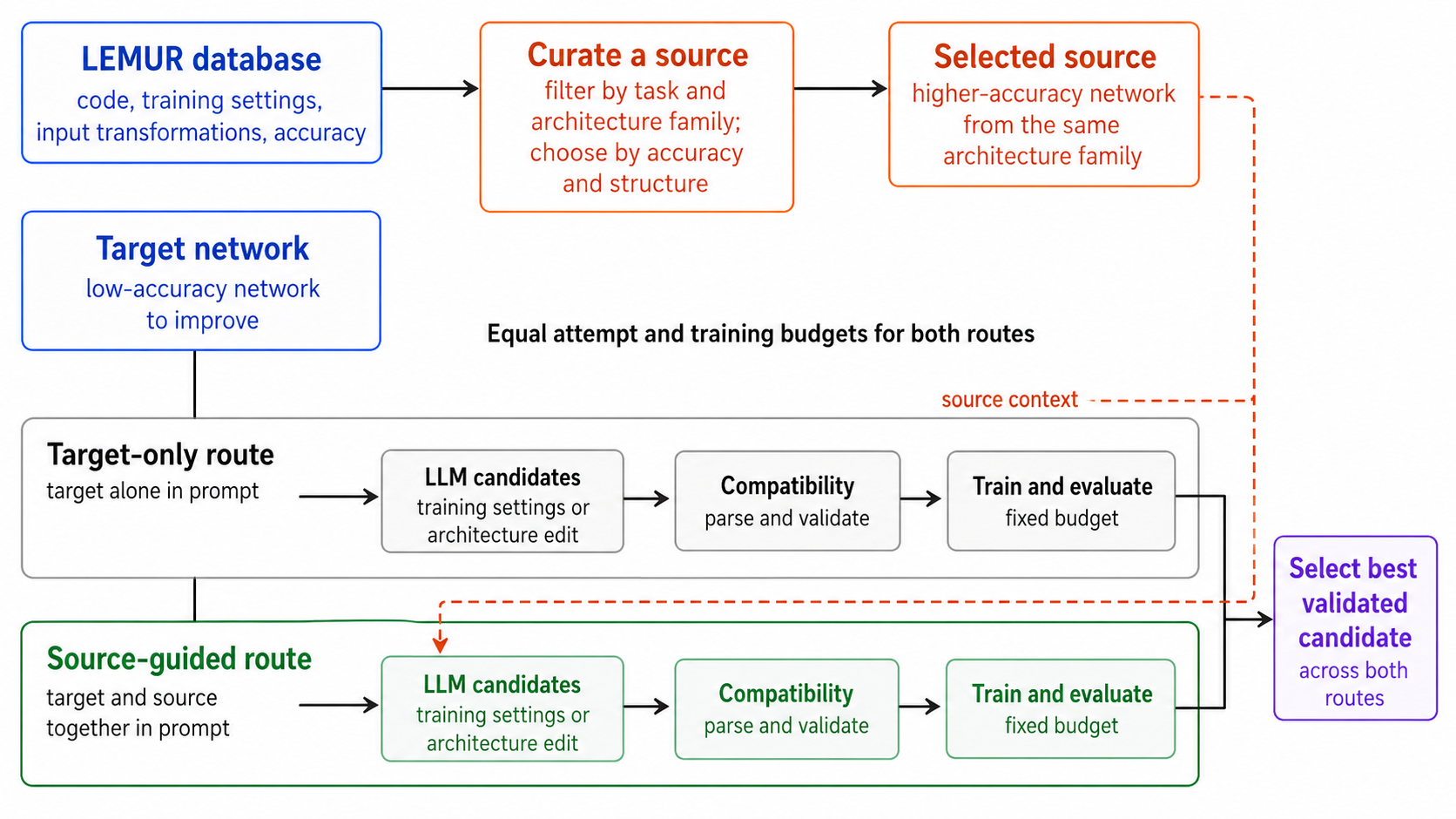}
\caption{Complete pipeline. A curated higher-accuracy network supplies context only to the source-guided route. The target-only and source-guided routes receive equal generation and training budgets, and the final step selects the strongest validated candidate across both routes.}
\label{fig:pipeline}
\end{figure}

\subsection{Types of Candidate Changes}

 For each target/source pair, the LLM generates candidate changes under four experimental conditions. Two conditions use information from the target alone, and two add information from the source. Within each information setting, one condition changes the training recipe and one allows a structured architecture edit. Every condition receives the same number of generation attempts, and every successful candidate is trained under the same evaluation budget. \cref{tab:action_spaces} names these four conditions for later reference.

\begin{table}[!ht]
\centering
\resulttablesize
\begin{tabular*}{\columnwidth}{@{\extracolsep{\fill}}lll@{}}
\toprule
Condition & Prompt contains & Candidate change \\
\midrule
\texttt{hp\_default} & Target only & Training settings \\
\texttt{baseline\_edit} & Target only & Architecture edit \\
\texttt{hp\_transfer} & Target and source & Training settings \\
\texttt{analogical\_edit} & Target and source & Architecture edit \\
\midrule
\texttt{hp\_copy} & --- (no LLM call) & Source recipe unchanged \\
\bottomrule
\end{tabular*}
\caption{Generation conditions used in the main experiments, plus the deterministic \texttt{hp\_copy} reference of \cref{sec:decomposition}. Training settings include hyperparameters and input transformations. Architecture-edit conditions may also adjust these settings.}
\label{tab:action_spaces}
\end{table}

The first two conditions use only the target, while the other two add the selected source. The fixed-architecture conditions generate training settings; the architecture-edit conditions may change the network structure as well.

In \texttt{hp\_transfer}, the target architecture remains fixed. The LLM receives the source training recipe and produces a candidate recipe for the target using the same output format as \texttt{hp\_default}. It may change the learning rate, momentum, batch size, dropout when supported, and input transformation. A deterministic compatibility step removes fields unsupported by the trainer and substitutes the dataset's standard transformation when a proposed transformation is incompatible. Candidate accuracy is measured only after generation and compatibility checking are complete.

A historical CIFAR-10 screening synthesis compared three output representations with DeepSeek-Coder-1.3B, one training epoch, and eight target-only plus eight source-guided attempts per representation. Complete-code responses produced 0/16 valid candidates, line-level patches 14/16, and predefined structured fields 16/16. The archived screen does not retain a stable target-record identifier and uses representation-specific prompts, so its accuracy values are reported in the Supplementary Material and are not directly compared with the fixed-target, $N=32$ main study. We therefore use constrained outputs in the main experiments. Both structured architecture edits and training-recipe changes remain in the four-condition study, while the later analysis focuses on recipe transfer because it produces the largest repeated improvements.

\subsection{Compatibility Checks}

Every generated candidate passes the same deterministic compatibility checks: output-format parsing, removal of unsupported hyperparameter fields, and verification of dataset-compatible input transformations. These checks make the generated output executable and compatible with the evaluator. Subsequent training and evaluation determine candidate quality.

\cref{fig:pipeline} summarizes the complete procedure. The pipeline reports successful evaluation rate, best accuracy, mean accuracy, improvement over the target, and the guided advantage. It returns the highest-accuracy candidate when that candidate improves upon the target.

\section{Experimental Setup}

\subsection{Metrics}

The experiments evaluate both the quality of the generated models and the effectiveness of the curated guidance. We report:

\begin{itemize}
    \item \textbf{valid evaluation count}: the number of generated candidates that successfully pass validation and training;
    \item \textbf{best valid accuracy}: the strongest candidate found under the fixed generation budget;
    \item \textbf{mean valid accuracy}: the average accuracy of successfully evaluated candidates;
    \item \textbf{gain over original}: improvement over the initial target network;
    \item \textbf{guided advantage}: best valid source-guided accuracy minus best valid target-only accuracy under equal generation budgets;
    \item \textbf{failure causes}: the recorded reasons candidates fail, summarized in \cref{sec:cross_dataset}.
\end{itemize}

Together these metrics measure whether curated examples improve reliability, search effectiveness, and consistency.

\subsection{Statistical Analysis}

Each accuracy comparison uses one dataset--LLM--target-network combination and, for CIFAR-10, one of two fixed source rules. The nine reported settings are not nine independent targets: six reuse the same CIFAR-10 target across two source rules and three LLMs, while three reuse the same SVHN target/source pair across the same LLMs. These settings are a fixed set rather than a random sample, so we read the direction and consistency of the paired outcomes alongside the interval. We compare the best candidate found with and without the higher-accuracy reference under equal attempt budgets. We also report mean candidate accuracy and successful-evaluation rates. Wilson 95\% confidence intervals summarize the evaluation rates. A paired bootstrap interval is reported only as a descriptive dispersion summary over these nine correlated settings; it is not interpreted as population-level evidence.

\subsection{Implementation}

Experiments use TuneNNGen and image-classification models from LEMUR. The LLM generates candidate architectures or training recipes from either the target network alone or the target network combined with a curated source network. All comparisons use the same generation and training budgets, while the LLM parameters remain fixed. Therefore, the experiments isolate the effect of the provided guidance example rather than changes in the optimizer or model.

\subsection{Reproducibility Protocol}

Two protocols appear below: the \emph{search protocol} (one epoch, scored on the available evaluation split) for the historical $N=32$ comparisons, and the \emph{validation-first protocol} (train-derived split, five epochs, three seeds) for the longer-training checks; captions state which applies. The available evaluation split is therefore part of the historical search protocol and is not treated as an untouched test set. Before each run, configuration files fix and record the dataset, LLM, target and source identifiers and accuracies, generation condition, compatibility mode, attempt budget, and seed offset. Main CIFAR-10 and SVHN AlexNet studies use $N=32$ attempts for each of the four conditions; smaller probes use $N=8$. Runs use one training epoch, temperature 0.75, top-$k=50$, top-$p=0.95$, and one prompt per generation call. The reproducibility record preserves the commands, software environment, model versions, hardware information, and instructions needed to repeat each run. Exact LEMUR record identifiers and protocol definitions are provided in the supplementary material. The ledger also records a historical CUDA adaptive-pooling nondeterminism issue; the strict evaluator used for the reported comparisons applies the corrected deterministic handling. Independent repeats test whether fresh generations preserve the direction and approximate size of the improvement.

The longer-training checks use a fixed validation split created from the training data. The CIFAR-10 follow-up ranks all distinct generated candidates on this split and trains the selected recipes for five epochs across three seeds. The CIFAR-100 and Imagenette checks freeze one target-only and one source-guided recipe from the historical one-epoch comparisons, before the validation re-evaluation, and retrain each recipe for five epochs across three seeds. Repeating these checks with the same seed reproduces the same final model file. None of these validation-first checks accesses the official test split.

\subsection{LLMs}

The current completed experiments include:
\begin{itemize}
    \item OlympicCoder-7B;
    \item DeepSeek-Coder-1.3B-Instruct (used in smaller exploratory runs);
    \item DeepSeek-Coder-6.7B-Instruct;
    \item Qwen2.5-Coder-7B-Instruct.
\end{itemize}

DeepSeek-Coder-6.7B and Qwen2.5-Coder-7B were added to test whether the observed effect also appears with other code LLMs that fit on a single RTX 4090.

\subsection{Datasets}

Experiments cover CIFAR-10, CIFAR-100, SVHN, Imagenette, and CelebA-Gender. The complete $N=32$ generation comparisons focus on CIFAR-10 and SVHN. Validation-first multi-seed checks on CIFAR-10, CIFAR-100, and Imagenette test whether selected recipe advantages persist with five training epochs. Additional exploratory probes broaden the evaluation across image sizes, class counts, task difficulty, and architecture families; their outcomes are summarized in \cref{sec:cross_dataset}.

\section{Results}

\subsection{Main Results}

The main experiments generate $N=32$ attempts under each of the four conditions. The target-only group combines \texttt{hp\_default} and \texttt{baseline\_edit}, giving 64 attempts for each dataset--LLM--source setting. The source-guided group combines \texttt{hp\_transfer} and \texttt{analogical\_edit}, giving the same 64-attempt budget. CIFAR-10 uses a low-accuracy target at 12.54\% and the two fixed source choices defined in \cref{sec:method}: \texttt{highsource} is the highest-accuracy eligible record, while \texttt{archbest} is the high-accuracy AlexNet-style record selected for its closer layer organization. SVHN uses a low-accuracy AlexNet target at 19.59\% and a higher-accuracy AlexNet source at 95.24\%.

\begin{table}[!ht]
\centering
\resulttablesize
\setlength{\tabcolsep}{2pt}
\resizebox{\columnwidth}{!}{%
\begin{tabular}{llrrrrrrr}
\toprule
LLM & Rule & T valid & T mean & T best & G valid & G mean & G best & Gain \\
\midrule
DeepSeek 6.7B & \texttt{archbest} & 60/64 & 16.73 & 23.98 & 64/64 & 43.66 & \textbf{50.49} & +26.51 \\
DeepSeek 6.7B & \texttt{highsource} & 62/64 & 16.84 & 23.54 & 63/64 & 32.66 & \textbf{42.13} & +18.59 \\
Qwen 7B & \texttt{archbest} & 30/64 & 16.66 & 21.56 & 36/64 & 39.66 & \textbf{47.57} & +26.01 \\
Qwen 7B & \texttt{highsource} & 26/64 & 16.62 & 23.86 & 45/64 & 33.63 & \textbf{48.14} & +24.28 \\
OlympicCoder 7B & \texttt{archbest} & 49/64 & 17.47 & \textbf{49.71} & 58/64 & 42.83 & 48.06 & -1.65 \\
OlympicCoder 7B & \texttt{highsource} & 42/64 & 16.84 & \textbf{48.76} & 56/64 & 33.83 & 40.33 & -8.43 \\
\bottomrule
\end{tabular}}
\caption{CIFAR-10 results with 32 attempts for each generation condition. T denotes target-only and G guided generation; mean is the average over successfully evaluated candidates. Accuracies are percentages and gains are percentage points; bold marks the higher best score. Each group combines two conditions, giving 64 attempts. Compact model labels denote DeepSeek-Coder-6.7B, Qwen2.5-Coder-7B, and OlympicCoder-7B. Search protocol.}
\label{tab:main_n32_cifar}
\end{table}

\begin{table}[!ht]
\centering
\resulttablesize
\setlength{\tabcolsep}{2pt}
\resizebox{\columnwidth}{!}{%
\begin{tabular}{lrrrrrrr}
\toprule
LLM & T valid & T mean & T best & G valid & G mean & G best & Gain \\
\midrule
DeepSeek 6.7B & 64/64 & 19.65 & 22.54 & 63/64 & 21.35 & \textbf{78.80} & +56.26 \\
Qwen 7B & 25/64 & 19.59 & 19.59 & 35/64 & 19.59 & 19.59 & 0.00 \\
OlympicCoder 7B & 45/64 & 19.62 & \textbf{20.93} & 44/64 & 19.40 & 19.65 & -1.28 \\
\bottomrule
\end{tabular}}
\caption{SVHN AlexNet results with 32 attempts for each generation condition. T denotes target-only and G guided generation; mean is the average over successfully evaluated candidates. Accuracies are percentages and gains are percentage points; bold marks the higher best score. Compact model labels are defined as in \cref{tab:main_n32_cifar}. Search protocol.}
\label{tab:main_n32_svhn}
\end{table}

\cref{tab:main_n32_cifar,tab:main_n32_svhn} present the central results. Across the nine main settings, the guided group achieves the highest best-candidate accuracy in five settings and ties in one. The SVHN majority-class baseline is 19.59\% (5,099 of 26,032 examples), so the Qwen guided result at 19.59\% is a degenerate majority-baseline outcome rather than evidence of a useful tie. Target-only generation gives the higher best accuracy in the three OlympicCoder rows. In the two CIFAR-10 OlympicCoder comparisons, target-only generation already finds candidates near 49\%, above the 10\% chance baseline. On SVHN, the OlympicCoder routes remain near the original 19.59\% target. Source guidance therefore does not add a stronger one-epoch recipe in these settings.

Across all nine settings, the mean guided advantage between the two best candidates is 15.59 percentage points, with a descriptive paired-bootstrap 95\% interval from 3.63 to 29.05 points. CIFAR-10 gives the clearest repeated pattern: guidance wins best accuracy in four of six settings and mean accuracy in all six. It also raises the successful-evaluation rate across the six CIFAR-10 rows from 70.1\% (Wilson 95\%: 65.3--74.4\%) to 83.9\% (79.8--87.2\%) and the weighted mean accuracy across evaluated candidates from 16.89\% to 37.80\%. Resampling both groups down to a matched number of evaluated candidates preserves the same four best-candidate wins; the resampling procedure and aggregate outcome are given in the supplement. Independent target-only replicates also vary: the best-candidate difference between two target-only runs is 0.44--2.30 points, while the valid-count difference is 2--7 candidates across the three LLMs.

Source curation changes the value of the guidance. \cref{tab:main_n32_cifar} compares the two fixed CIFAR-10 source choices. \texttt{archbest} gives a larger guided advantage than \texttt{highsource} for DeepSeek-Coder-6.7B (26.51 versus 18.59 points) and Qwen2.5-Coder-7B (26.01 versus 24.28 points). It also improves the OlympicCoder-7B comparison from $-8.43$ to $-1.65$ points. The architecture-related source therefore produces the more favorable guided advantage for all three LLMs, though both OlympicCoder differences remain negative; the comparison uses one target and two fixed records. The strongest one-epoch result occurs on SVHN AlexNet, where \texttt{hp\_transfer} reaches 78.80\% compared with 22.54\%, a gain of 56.26 points.

\subsection{Repeats and Longer-Training Checks}

Fresh generations preserve the two strongest early-training results. The CIFAR-10/DeepSeek-Coder-6.7B \texttt{archbest} advantage is 26.51, 26.75, and 28.03 percentage points across three runs. The SVHN AlexNet/DeepSeek-Coder-6.7B advantage is 56.26 points in the original run and 55.59 points in an independent repeat. Both comparisons use the same 32-attempt budget for every generation condition.

The one-epoch experiments provide a fast candidate search. We then freeze selected recipes and train them for five epochs across three seeds. \cref{tab:one_vs_five} places the early and longer-training scores for those exact recipes side by side.

\begin{table}[!ht]
\centering
\resulttablesize
\setlength{\tabcolsep}{3pt}
\resizebox{\columnwidth}{!}{%
\begin{tabular}{lllrrrrrr}
\toprule
& & & \multicolumn{3}{c}{One epoch (\%)} & \multicolumn{3}{c}{Five epochs (\%)} \\
\cmidrule(lr){4-6}\cmidrule(lr){7-9}
Dataset & LLM & Arch. & T & G & Gain & T & G & Gain \\
\midrule
CIFAR-10 & DeepSeek 6.7B & AlexNet-style & 22.51 & 50.54 & +28.03 & 35.59 & \textbf{76.53} & \textbf{+40.94} \\
Imagenette & DeepSeek-Coder-1.3B & AlexNet & 12.41 & 30.85 & +18.45 & 43.86 & \textbf{62.69} & \textbf{+18.83} \\
CIFAR-100 & OlympicCoder-7B & AlexNet & 1.31 & 10.52 & +9.21 & 18.46 & \textbf{25.73} & \textbf{+7.27} \\
SVHN & DeepSeek-Coder-6.7B & AlexNet & 19.52 & \textbf{76.51} & \textbf{+56.99} & 89.35 & 89.62 & +0.27 \\
\bottomrule
\end{tabular}}
\caption{Selected fixed-recipe comparisons, ordered by five-epoch gain. T uses a selected target-only recipe and G a selected guided recipe. The one-epoch values are historical available-split scores of the frozen recipes and differ from the group maxima in \crefrange{tab:main_n32_cifar}{tab:main_n32_svhn}; five-epoch values are three-seed means on train-derived validation splits. The Imagenette pilot used DeepSeek-Coder-1.3B with one attempt per condition. On SVHN, 19.52 is the frozen target-only recipe score; 19.59 is the stored-target baseline and 22.54 the separate main-search maximum.}
\label{tab:one_vs_five}
\end{table}

The CIFAR-10 follow-up uses the third independent \texttt{archbest} run. Each generation condition again receives 32 attempts, giving equal 64-attempt budgets to the target-only and guided groups. Among 121 successfully evaluated outputs, the target-only group produces one distinct recipe and the guided group produces four. Thus, the five-epoch comparison is a selected-recipe check rather than an estimate over 121 independent architectures. After ranking the five recipes on the validation split, we train each one for five epochs over three seeds.

Across three training runs, the best guided recipe averages 76.53\% accuracy compared with 35.59\% for the target-only recipe, a gain of \textbf{40.94 percentage points}. The guided recipe wins in all three runs. Selection and training use the train-derived validation split throughout this follow-up.

The cross-dataset checks extend the persistent gain beyond CIFAR-10. On Imagenette, the guided recipe reaches $62.69\pm0.70$\% compared with $43.86\pm1.20$\%, a gain of \textbf{18.83 points}. This frozen pair (12.41\% target-only and 30.85\% guided at one epoch) comes from a strict one-attempt-per-condition pilot. A separate direct-copy decomposition in the Supplementary Material uses the same stored target but a different historical portfolio, where 23.26\% is the original record and 29.40\%/36.56\% are portfolio maxima; the gains are therefore not expected to match. On CIFAR-100, the guided recipe reaches $25.73\pm0.51$\% compared with $18.46\pm0.51$\%, a gain of \textbf{7.27 points}. The guided recipe wins all three seeds on both datasets. On SVHN, guidance finds a much faster one-epoch recipe while the target-only recipe catches up by five epochs, placing the value of guidance in the budget-constrained regime. All longer-training checks use train-derived validation splits.

\subsection{A Concrete CIFAR-10 Recipe Example}

\cref{tab:cifar_recipe_example} shows the exact winning recipes from the third CIFAR-10 run. Both candidates use the same network architecture, learning rate (0.0480), momentum (0.5178), and \texttt{norm\_256\_flip} input transformation. The guided route finds a smaller batch size and lower dropout. On the 45,000-image training split, batch size 64 gives approximately 704 optimizer updates per epoch, compared with 88 updates for batch size 512. This is a plausible mechanism for the early difference, but the study does not include a causal equal-update or batch-size-only ablation; the result should therefore not be attributed to batch size alone.

\begin{table}[!ht]
\centering
\resulttablesize
\begin{tabular*}{\columnwidth}{@{\extracolsep{\fill}}lrrrr@{}}
\toprule
Generation route & Batch & Dropout & One epoch & Five epochs \\
\midrule
Target only & 512 & 0.2939 & 22.51\% & 35.59\% \\
Same-family guidance & 64 & 0.2000 & \textbf{50.54\%} & \textbf{76.53\%} \\
\bottomrule
\end{tabular*}
\caption{Exact CIFAR-10 winning recipes from the third equal-budget run. Five-epoch values are validation means across three runs of the same frozen recipes. Validation-first protocol.}
\label{tab:cifar_recipe_example}
\end{table}

The guided recipe improves one-epoch accuracy by 28.03 points in this run and five-epoch mean accuracy by 40.94 points. Across three generations of the same experiment, the one-epoch advantages are 26.51, 26.75, and 28.03 points. In this run, the winning guided recipe combines the smaller source batch size with the target's learning rate, momentum, and transformation. The recipes differ mainly in batch size, but the advantage persists at five epochs. This observation motivates a controlled update-count ablation rather than proving that batch size is the causal factor.

\subsection{Cross-Dataset Results and Reliability}
\label{sec:cross_dataset}

Additional experiments test how the method behaves across architecture families and datasets. The new five-epoch, three-seed checks show persistent source-guided gains on CIFAR-100 and Imagenette AlexNet, adding 7.27 and 18.83 percentage points, respectively. Earlier one-epoch Imagenette AlexNet comparisons also give three source-guided wins ranging from $+1.81$ to $+7.16$ points. Both groups reach approximately 96\% on CelebA-Gender; because a split-specific majority baseline was not recorded, this result is reported descriptively and is not used as evidence for source guidance.

Smaller SVHN tests on BagNet, AirNext, and DPN68 produce mixed results (the per-architecture values are reported in the supplement). Overall, low-accuracy AlexNet-style targets on CIFAR-10, CIFAR-100, SVHN, and Imagenette form the strongest current application regime.

The validity analysis records why generated candidates fail before or during evaluation. Observed causes include code-extraction errors, spatial-dimension collapse in small-image models, unsupported hyperparameter fields, accuracy below the evaluator threshold on Imagenette, and training timeouts on a larger-image dataset. These outcomes are included in the reported generation and evaluation success rates.

\subsection{How the Source Recipe Contributes}
\label{sec:decomposition}

The strongest source-guided condition in the main experiments is \texttt{hp\_transfer}, where the LLM uses the source training settings to propose a new recipe while keeping the target architecture fixed. To measure how useful the recorded source recipe is by itself, we add the deterministic \texttt{hp\_copy} condition of \cref{tab:action_spaces}: it applies the source recipe to the target with no LLM call, giving one candidate per pair and consuming no generation budget.

The four-setting decomposition is reported in the Supplementary Material. For clarity, its contribution values are measured relative to the best target-only candidate in the same archived portfolio, not the stored target record; the stored-target comparisons below are separate. Direct copying reaches 42.67\% on CIFAR-10 \texttt{archbest} and 35.06\% on \texttt{highsource}, while the corresponding LLM-guided recipes reach 50.49\% and 42.13\%. The other two settings are SVHN (19.59\% direct copy, 78.80\% guided) and Imagenette (28.92\% direct copy, 36.56\% guided).

SVHN AlexNet demonstrates the value of adapting the recorded settings. Directly copying the source recipe gives 19.59\%, equal to the original target accuracy. The copied recipe uses learning rate $6.747\times10^{-5}$, momentum $0.980$, batch size 256, and transformation \texttt{norm\_299\_flip}. The learning rate comes from a network already near 95\% accuracy and produces little target training progress in one epoch. With the same source information, \texttt{hp\_transfer} generates a different recipe that reaches 78.80\%. The 59.21-point difference shows that the generated recipe is substantially more effective for the low-accuracy target than literal copying.

Imagenette AlexNet provides another example. Direct recipe copy improves the target from 23.26\% to 28.92\%, the best target-only candidate reaches 29.40\%, and the LLM-generated source-guided recipe reaches 36.56\%. Relative to the best target-only candidate in each archived portfolio, the gain therefore splits into what copying achieves and what the LLM adds: $+18.69/+7.82$ on CIFAR-10 \texttt{archbest}, $+11.52/+7.07$ on \texttt{highsource}, $-2.95/+59.21$ on SVHN, and $-0.48/+7.64$ on Imagenette, each pair summing to the reported gain. Copying helps when source and target sit at similar points in training; the LLM contribution is positive in all four settings.

\section{Discussion}

\paragraph{Source selection is data curation.}
The selected guiding network determines which prior information is available to the LLM: architecture code, training settings, input transformations, and measured performance. Therefore, selecting a source network is a data-curation decision. On CIFAR-10, the architecture-related \texttt{archbest} source produces a more favorable guided advantage than the highest-accuracy \texttt{highsource} source for all three evaluated LLMs, even though its stored accuracy is 7.67 points lower. This shows that the most accurate stored network is not necessarily the most useful example for improving another network. Similarity and transferability provide complementary criteria for selecting useful guidance.

\paragraph{Understanding what transfers.}
The recipe-copy ablation separates the contribution of the selected network itself from the additional adaptation performed by the LLM. A source recipe can directly improve a target when its training settings transfer effectively, but the LLM can also modify the recorded settings into a more suitable target-specific recipe. These two effects behave differently across datasets. For example, on CIFAR-10 the copied recipes already provide substantial gains, while on SVHN the copied high-performing recipe leaves the target unchanged and the LLM-generated adaptation provides the full improvement. Thus, stored accuracy describes how well a recipe performed for the network that produced it, but it does not directly measure how useful that experiment is as guidance for another network.

\section{Limitations}

The positive results concentrate in AlexNet-style targets on CIFAR-10, CIFAR-100, SVHN, and Imagenette; other architecture families are small, mixed, or use smaller budgets. The main search uses one epoch, 32 attempts per condition, and the available evaluation split, while selected five-epoch checks use a train-derived validation split and do not cover every generated candidate or provide an untouched-test confirmation for every headline setting. Source guidance can produce more valid and more distinct recipes, changing the effective best-of-$N$ opportunity; matched-count resampling and raw successfully evaluated-candidate distributions in the Supplementary Material show that the four positive CIFAR-10 rows remain positive in every descriptive draw, but they do not remove dependence between generated candidates. The historical five-epoch \texttt{highsource} Olympic7B comparison is an explicit boundary case: target-only reaches 79.93\% while guided reaches 38.89\%, a $-41.04$-point reversal. Longer-training evidence covers selected recipe pairs, the four-setting \texttt{hp\_copy} ablation is not a complete causal decomposition, and the batch-size explanation has no equal-update ablation. The additional HPO evidence is a 16-recipe validation-only Sobol sweep rather than a completed 64-trial Optuna comparison. The source-quality stress test finds no reliable monotonic benefit: 22 of 24 raw outputs and final recipes are identical across source levels, although prompt and transform choices remain confounded; a matched held-out replay also found the existing target-only LLM recipe above source conditioning in one setting. Detailed evidence for these validation-only, source-quality, and held-out analyses is provided in the Supplementary Material. Historical training is stochastic, with independent target-only replicates varying by 0.44--2.30 points in best accuracy.

\section{Conclusion}

This paper studies neural-network curation for equal-budget LLM-guided model improvement. Curated guidance can improve candidate quality through recipe transfer, but its value depends on the target/source pair: the architecture-related CIFAR-10 source is more favorable than the highest-accuracy source in this controlled comparison, while held-out and source-quality controls prevent a universal claim.

\paragraph{Artifact availability.}
The supplementary package provides exact LEMUR record identifiers, protocol definitions, stored evidence tables, and deterministic analysis scripts used for the reported comparisons. The cleaned full implementation is not part of this release.

\vspace{0.4cm}
\noindent\textbf{Acknowledgments.}
This work was partially supported by the Alexander von Humboldt Foundation.

{
\small
\bibliographystyle{splncs04}
\bibliography{references}
}

\end{document}


\maketitle

\section{Scope and artifact statement}

This supplement documents the protocols, identifiers, stored result summaries, and deterministic analyses used for the camera-ready paper. It is intended to make the reported comparisons auditable without implying that every historical experiment belongs to one pooled statistical sample. Historical one-epoch search results, validation-first longer-training checks, validation-only exploratory runs, and matched held-out replays are kept as separate protocol classes.

The supplementary material provides the exact headline prompt JSON files, LEMUR record identifiers, protocol definitions, stored result summaries, candidate-level reanalysis, and deterministic analysis scripts used in the reported comparisons. The cleaned full implementation is not part of this release.

\section{Exact target and source records}

\begin{longtable}{p{0.25\linewidth}p{0.45\linewidth}r}
\toprule
Role & LEMUR record identifier & Stored accuracy \\
\midrule
\endhead
CIFAR-10 target & \code{ed1246fb7fb1bb50c9127eebcb6e05d2} & 0.1254 \\
CIFAR-10 \code{archbest} & \code{ed44f284382e3593f982eeaa71996065} & 0.7427 \\
CIFAR-10 \code{highsource} & \code{f6923a5f94145abd0a362581ae0012e9} & 0.8194 \\
SVHN target & \code{82f102e1bd4884ac574a1543d9eac6fb} & 0.1959 \\
SVHN source & \code{9da54aec32cd2aebd2374d086521e20f} & 0.9524 \\
CIFAR-100 historical model & \code{edit-b76f0bbaf929a754475beed0e6c66065} & -- \\
Imagenette historical model & \code{edit-0402832feb25824ef4924e28f374d38f} & -- \\
\bottomrule
\end{longtable}
\suppcaption{Exact target and source records used by the headline comparisons.}

The CIFAR-10 source-selection comparison is fixed before generation. \code{highsource} is the highest-accuracy eligible record; \code{archbest} is selected by the manual AlexNet-style layer criterion described in the paper. No learned source retriever is used in the headline comparison.

The internal registry label \code{alt\_nn1} denotes the AlexNet-style CIFAR-10 target used in the selected-recipe table of the main paper.

\section{Candidate conditions and protocol}

\begin{table}[ht]
\centering
\begin{tabularx}{\linewidth}{p{0.22\linewidth}p{0.27\linewidth}Y}
\toprule
Condition & Prompt information & Change allowed \\
\midrule
\code{hp\_default} & Target only & LLM changes training settings while the target architecture remains fixed. \\
\code{hp\_transfer} & Target plus source & LLM adapts source training settings to the fixed target architecture. \\
\code{baseline\_edit} & Target only & Structured architecture-edit condition. \\
\code{analogical\_edit} & Target plus source & Structured architecture-edit condition using the source as context. \\
\code{hp\_copy} & Target plus source, no LLM call & Directly copies the source recipe to the target as a deterministic reference. \\
\bottomrule
\end{tabularx}
\suppcaption{Candidate conditions. Preliminary output-format screening is not a main experimental arm.}
\end{table}

\section{Exact headline prompt configurations}

The four headline CIFAR-10 routes use the exact JSON configurations archived under \code{artifacts/prompts/}. The JSON files are the authoritative, byte-preserved configurations; the table records the resolved prompt identifier and the fields that distinguish the routes.

\begin{table}[ht]
\centering
\scriptsize
\begin{tabularx}{\linewidth}{p{0.19\linewidth}p{0.37\linewidth}Y}
\toprule
Condition & Exact configuration and prompt identifier & Resolved generation contract \\
\midrule
\code{hp\_default} & \url{NN_gen_frozen_image_starter.json}; \url{improve_classification_only_cifar10_frozen_hp_default} & Target blocks only; fixed architecture; hyperparameters and transform may change; XML \code{hp}, \code{tr}, \code{edit} output. \\
\code{hp\_transfer} & \url{NN_gen_frozen_image_starter.json}; \url{improve_classification_only_cifar10_frozen_hp_transfer} & Target plus source blocks; fixed architecture; source recipe and transform are the main transfer; XML \code{hp}, \code{tr}, \code{edit} output. \\
\code{baseline\_edit} & \url{NN_gen_cifar10_edit_paper.json}; \url{improve_classification_only_cifar10_edit_paper} & Target blocks only; safe structured CIFAR edit or \code{hp\_transform\_only}; XML \code{hp}, \code{tr}, \code{edit} output. \\
\code{analogical\_edit} & \url{NN_gen_analogical_cifar10_edit_paper.json}; \url{improve_classification_only_analogical_cifar10_edit_paper} & Target plus source edit hint; transfer an improvement pattern through the safe structured-edit schema; XML \code{hp}, \code{tr}, \code{edit} output. \\
\bottomrule
\end{tabularx}
\suppcaption{Exact generation configurations for the main CIFAR-10 runs. All four configurations use the target fields \code{nn\_code}, \code{accuracy}, \code{metric}, \code{transform\_code}, and \code{prm}; guided routes additionally receive the source fields specified in their JSON file.}
\end{table}

The frozen output contract is reproduced below. Source guidance changes the context, not the three-tag schema. The guided prompt asks the model to adapt the source recipe.

\begin{verbatim}
<hp>
{"batch":64,"transform":"echo_32","lr":0.01,"momentum":0.9}
</hp>
<tr>
import torchvision.transforms as transforms
def transform(_):
    return transforms.Compose([transforms.Resize((32, 32)),
                               transforms.ToTensor()])
</tr>
<edit>
{"mode":"hp_transform_only"}
</edit>
\end{verbatim}

The two structured-edit JSON files preserve the same three-tag output schema but allow only the documented safe keys: \code{mode}, \code{width}, \code{pool}, \code{residual\_kernel}, and \code{classifier\_dropout}. The analogical file additionally supplies the source edit, source accuracy, and source recipe as an add-on context. Shared generation settings are temperature 0.75, top-$k=50$, top-$p=0.95$, one prompt per call, and 32 attempts per condition.

\subsection{Concrete source-record serialization}

The following is a concrete serialization example for the CIFAR-10 \code{archbest} source record. It shows the fields supplied to the prompt; the full architecture and transform strings remain in the LEMUR-backed record and the exact prompt JSON uses the corresponding placeholders.

\begin{verbatim}
<source_record>
  <record_id>ed44f284382e3593f982eeaa71996065</record_id>
  <stored_accuracy>0.7427</stored_accuracy>
  <source_hp>
    {"batch":64,"dropout":0.4920805165236665,
     "lr":0.1665845971036694,"momentum":0.031072583837780998,
     "transform":"norm_299"}
  </source_hp>
  <source_tr>transform block for norm_299</source_tr>
  <source_nn>architecture block, nn_id=
    08a5123a16c88616ce01d3c8167cb6fe</source_nn>
  <output_schema>
    exactly one <hp>, one <tr>, and one <edit> tag;
    no prose or unified diff
  </output_schema>
</source_record>
\end{verbatim}

The target record in this pair is \code{ed1246fb7fb1bb50c9127eebcb6e05d2} with stored accuracy 0.1254. The serialization distinguishes the record identifier, stored score, recipe, transform block, architecture block, and output schema rather than presenting the source as an undifferentiated text example.

\subsection{Preliminary output-format screening}

This historical CIFAR-10 screening synthesis compares formatting strategies using DeepSeek-Coder-1.3B, one training epoch, and eight target-only plus eight source-guided attempts per representation. It is reported for reproducibility only and is not a main experimental arm. The archived screen does not retain a stable target-record identifier, and its representation-specific prompts were collected across historical screening runs; its accuracies are therefore diagnostics rather than values directly comparable with the fixed-target, $N=32$ main study. A full-code response produced no parseable candidates; patch and structured-field responses were evaluated under their respective parsers.

\begin{table}[ht]
\centering
\begin{tabular}{lrr}
\toprule
Output format & Valid outputs & Best accuracy \\
\midrule
Full Python code & 0/16 & -- \\
Patch & 14/16 & 55.57\% \\
Structured fields & 16/16 & 62.82\% \\
\bottomrule
\end{tabular}
\suppcaption{Historical output-format screening synthesis: DeepSeek-Coder-1.3B, CIFAR-10, one epoch, and 8 target-only plus 8 source-guided attempts per representation. The screen motivates the structured XML/field contract but is not a controlled main-study comparison.}
\end{table}

The historical search protocol uses one training epoch, 32 generation attempts per condition, temperature 0.75, top-$k=50$, top-$p=0.95$, and one prompt per generation call. Target-only and source-guided groups each combine two conditions and therefore receive 64 attempts per setting. The historical search score uses the available evaluation split. The validation-first protocol instead creates a fixed split from the training data, selects or freezes recipes on that split, and trains selected recipes for five epochs over three seeds.

\paragraph{Algorithm S1: equal-budget comparison.}
For each fixed target/source pair: (1) write a protocol record containing the dataset, record identifiers, LLM, condition, attempt budget, and seed offset; (2) generate exactly 32 candidate outputs for each condition; (3) apply the strict parser and evaluator; (4) retain valid accuracy, failure cause, and canonicalized recipe fields; (5) aggregate each target-only pair and source-guided pair separately; (6) report valid count, best valid accuracy, mean valid accuracy, and the guided advantage. Candidate failures remain in the denominator of the valid-count metric. The evaluator rejects structurally invalid candidates before GPU execution.

\section{Full historical one- and five-epoch table}

The following table preserves the eight-row historical result file used to check the selected fixed-recipe table in the paper. These rows use the historical protocol and should not be interpreted as if every recipe was selected with the later validation-first split.

\begin{longtable}{p{0.12\linewidth}p{0.18\linewidth}p{0.16\linewidth}rrrrr}
\toprule
Dataset & LLM & Source rule & T-1e & G-1e & T-5e & G-5e & $\Delta$-5e \\
\midrule
\endhead
CIFAR-10 & DS6.7B & \code{archbest} & 23.98 & 50.49 & 48.39 & 76.86 & +28.47 \\
CIFAR-10 & Qwen2.5-7B & \code{archbest} & 21.56 & 47.57 & 49.93 & 73.10 & +23.17 \\
CIFAR-10 & Olympic7B & \code{archbest} & 49.71 & 48.06 & 78.80 & 77.80 & -1.00 \\
CIFAR-10 & Olympic7B & \code{highsource} & 48.76 & 40.33 & 79.93 & 38.89 & -41.04 \\
SVHN & DS6.7B & AlexNet & 22.54 & 78.80 & 88.52 & 89.85 & +1.33 \\
SVHN & Qwen2.5-7B & AlexNet & 19.59 & 19.59 & 85.86 & 88.38 & +2.52 \\
SVHN & Olympic7B & AlexNet & 20.93 & 19.65 & 88.99 & 88.50 & -0.50 \\
SVHN & DS6.7B & AlexNet repeat 2 & 25.09 & 80.69 & 88.37 & 90.47 & +2.10 \\
\bottomrule
\end{longtable}
\suppcaption{Full historical one- and five-epoch values used to check the selected fixed-recipe comparisons.}

The values in this table are percentages. For example, the CIFAR-10 DeepSeek row has a five-epoch guided advantage of $76.86-48.39=28.47$ points. Using the underlying unrounded records, the separate selected-recipe table in the paper reports $25.73-18.46=7.27$ for CIFAR-100 and $89.62-89.35=0.27$ for SVHN.

The Olympic7B \code{highsource} row is an important boundary case: the target-only recipe reaches 79.93\% after five epochs, whereas the source-guided recipe reaches 38.89\%, a $-41.04$-point reversal. It shows why the positive one-epoch patterns cannot be promoted to a universal claim about source guidance.

\section{Additional reported result tables}

\subsection{Direct-copy decomposition}

\begin{table}[ht]
\centering
\begin{tabular}{lrrrr}
\toprule
Setting & Original & Target only & Direct copy & LLM guided \\
\midrule
CIFAR-10 \code{archbest} & 12.54\% & 23.98\% & 42.67\% & \textbf{50.49\%} \\
CIFAR-10 \code{highsource} & 12.54\% & 23.54\% & 35.06\% & \textbf{42.13\%} \\
SVHN AlexNet & 19.59\% & 22.54\% & 19.59\% & \textbf{78.80\%} \\
Imagenette AlexNet & 23.26\% & 29.40\% & 28.92\% & \textbf{36.56\%} \\
\bottomrule
\end{tabular}
\suppcaption{One-epoch accuracy when the source recipe is copied directly or used by the LLM to generate an adapted recipe. The copied recipe is fixed for each target/source pair. The Imagenette row is a separate historical portfolio on the same stored target as the strict one-attempt-per-condition frozen pair in main-paper Table~4; 23.26\% is the stored target accuracy, while 29.40\% and 36.56\% are portfolio maxima. Search protocol.}
\label{tab:supp_hp_copy_decomp}
\end{table}

\subsection{SVHN architecture-family probes}

\begin{longtable}{p{0.18\linewidth}p{0.18\linewidth}r rrrr}
\toprule
Architecture & LLM & Valid T/G & Best T & Best G & Guided advantage \\
\midrule
\endhead
AirNext & DS6.7B & 16/10 & 76.78\% & 74.64\% & -2.15 \\
AirNext & Olympic7B & 15/11 & 74.99\% & 76.54\% & +1.55 \\
AirNext & Qwen2.5-7B & 4/8 & 74.91\% & 74.73\% & -0.18 \\
BagNet & DS6.7B & 16/16 & 83.77\% & 82.99\% & -0.78 \\
BagNet & Olympic7B & 14/14 & 83.90\% & 83.29\% & -0.62 \\
BagNet & Qwen2.5-7B & 5/10 & 83.27\% & 83.03\% & -0.24 \\
DPN68 & DS6.7B & 16/16 & 71.92\% & 68.84\% & -3.08 \\
DPN68 & Olympic7B & 14/13 & 79.45\% & 73.65\% & -5.80 \\
DPN68 & Qwen2.5-7B & 7/9 & 72.58\% & 70.65\% & -1.93 \\
\bottomrule
\end{longtable}
\suppcaption{SVHN architecture-family probes. These are smaller robustness probes rather than headline comparisons.}

Each route pools two conditions with eight attempts each, giving 16 attempts per route. These are smaller $N=8$ robustness probes rather than headline comparisons; unequal valid counts are shown explicitly and the best scores are percentages.

\section{Reproducibility and independent variation}

The third CIFAR-10 follow-up has 121 successfully evaluated outputs: one distinct successful target-only recipe and four distinct successful source-guided recipes. Those five recipes were ranked on a train-derived validation split and then trained for three seeds. The best target-only and guided recipes have five-epoch means 35.59\% and 76.53\%, respectively. The independent target-only replicate checks show a 0.44--2.30 point spread in best accuracy and a 2--7 candidate spread in valid counts across the three LLMs. These values provide a noise floor for small best-of-$N$ differences.

For the equal-count reanalysis, each target-only and source-guided comparison is reduced to $k=\min(n_T,n_G)$ valid candidates. Sampling is without replacement within each group, using a fixed seed; the best score is computed for each resample and the guided advantage is retained. The stored audit reports that the four CIFAR-10 guided best-candidate wins remain after this matched-count reduction. The analysis is descriptive because the valid candidate lists are generated by correlated prompts and the groups can have different recipe duplication rates.

\begin{table}[ht]
\centering
\scriptsize
\begin{tabular}{lccrrrrr}
\toprule
Setting & $n_T/n_G$ & $k$ & \shortstack{Mean best-$k$\\T} & \shortstack{Mean best-$k$\\G} & \shortstack{Mean best-$k$\\$\Delta$} & Descriptive 2.5--97.5\% & $P(\mathrm{G}>\mathrm{T})$ \\
\midrule
CIFAR-10 archbest DS & 60/64 & 60 & 23.98 & 50.48 & +26.50 & [26.35, 26.51] & 100\% \\
CIFAR-10 archbest Olympic & 49/58 & 49 & 49.71 & 47.92 & $-1.79$ & [$-2.55$, $-1.65$] & 0\% \\
CIFAR-10 archbest Qwen & 30/36 & 30 & 21.56 & 47.46 & +25.90 & [24.69, 26.01] & 100\% \\
CIFAR-10 highsource DS & 62/63 & 62 & 23.54 & 42.11 & +18.57 & [18.59, 18.59] & 100\% \\
CIFAR-10 highsource Olympic & 42/56 & 42 & 48.76 & 40.25 & $-8.51$ & [$-9.54$, $-8.43$] & 0\% \\
CIFAR-10 highsource Qwen & 26/45 & 26 & 23.86 & 46.40 & +22.54 & [14.73, 24.28] & 100\% \\
SVHN AlexNet DS & 64/63 & 63 & 22.50 & 78.80 & +56.30 & [56.26, 56.26] & 100\% \\
SVHN AlexNet Olympic & 45/44 & 44 & 20.90 & 19.65 & $-1.26$ & [$-1.28$, $-1.28$] & 2.08\% \\
SVHN AlexNet Qwen & 25/35 & 25 & 19.59 & 19.59 & 0.00 & [0.00, 0.00] & 0\% \\
\bottomrule
\end{tabular}
\suppcaption{Matched-count reanalysis over successfully evaluated candidates. Mean best-$k$ T/G are Monte Carlo averages of the resampled best-of-$k$ maxima, and mean best-$k$ $\Delta$ is their difference; these are not the paper's mean-valid-candidate-accuracy metric. Values are percentages. Each row uses 10,000 without-replacement draws with seed 20260810; quantiles are descriptive resampling quantiles, not population confidence intervals. The four positive CIFAR-10 rows are positive in every draw, both Olympic7B CIFAR-10 rows are negative in every draw, and the SVHN rows retain the strong DeepSeek gain, majority-baseline Qwen tie, and mostly negative Olympic7B result.}
\end{table}

The historical ledger also records a CUDA adaptive-pooling nondeterminism issue. The strict evaluator used for the reported comparisons applies the corrected deterministic handling. The artifact archive contains the non-sensitive evidence files listed in Section 7; raw launch manifests and candidate-level records are not part of the release.

Figure S1 makes the best-of-budget interpretation explicit. In the SVHN/DeepSeek guided route, 57 of 63 successfully evaluated candidates equal the 19.59\% majority-class baseline and only two exceed 30\%; the isolated orange star is the 78.80\% route maximum rather than a shift of the whole candidate distribution.

\begin{figure}[p]
\centering
\includegraphics[width=\linewidth]{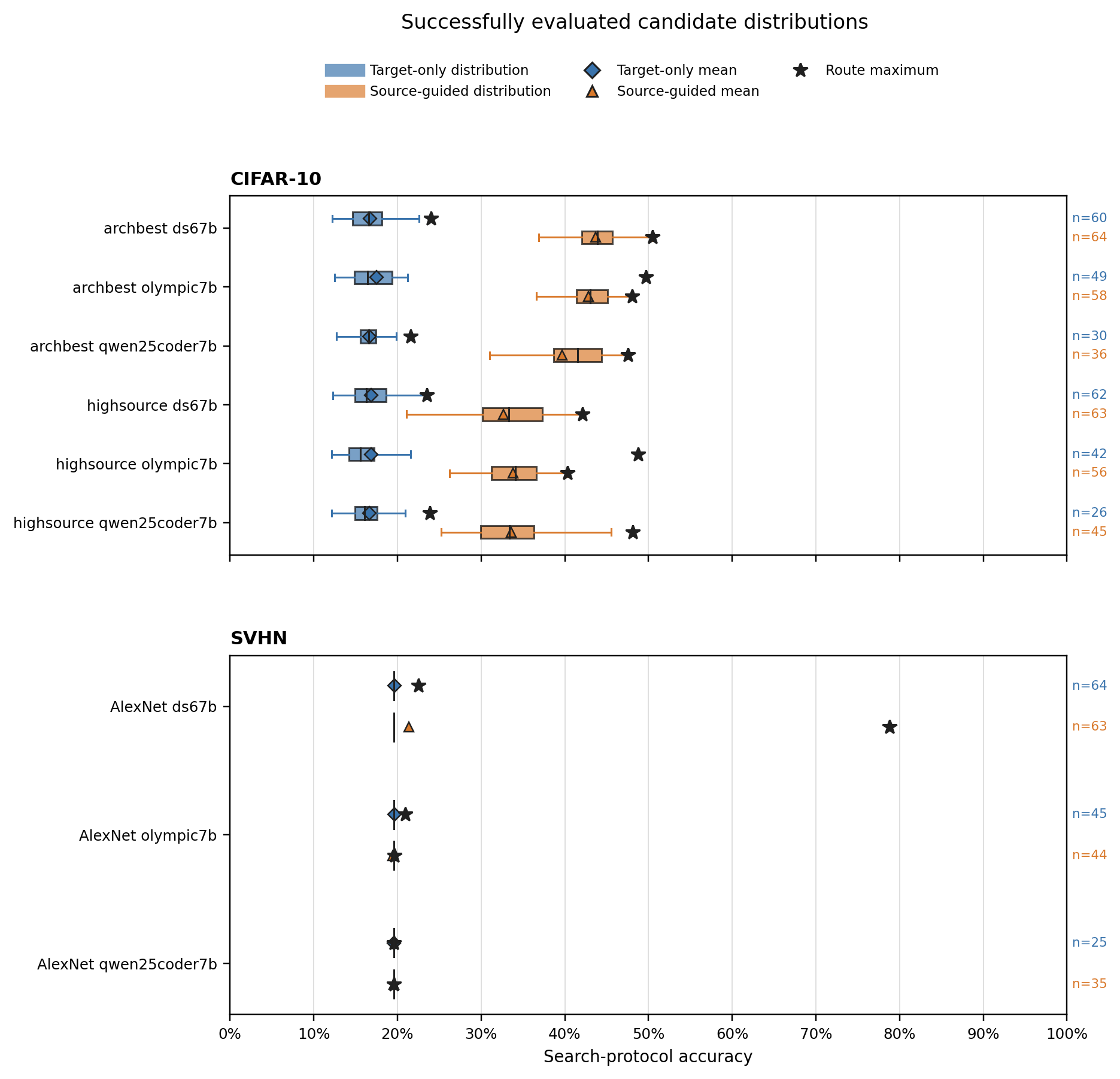}
\suppfigcaption{Successfully evaluated candidate distributions for the six CIFAR-10 and three SVHN settings. Blue boxes show target-only distributions and orange boxes show source-guided distributions; blue diamonds and orange triangles mark the corresponding route means, while black stars mark route maxima. Counts are printed beside each route. The boxes and markers summarize the successfully evaluated candidates shown by the historical search protocol; the figure is not the matched-count resampling display of Table S8.}
\end{figure}
\clearpage

\section{Stored validation evidence}

\subsection{Validation-only recipe search}

An overnight validation-only exploration completed 96/96 HPO arms: 16 recipes, two targets, and three seeds. The source-independent Sobol search selected \code{hpo\_11}, with batch size 32, dropout 0.058351, learning rate 0.0111643991, and momentum 0.731761. In the table, \emph{control} is the unchanged target baseline; \emph{direct copy} applies the source recipe without an LLM call; \emph{existing target-only recipe} is the previously selected target-only LLM recipe; and \emph{Sobol HPO winner} is the best of the 16 source-independent Sobol recipes.

The \emph{contrastive-prompt} route is a validation-only exploratory portfolio for the weak CIFAR-10 target, separate from the five main conditions in Table S2. It contains 16 generated candidates spanning aggressive, conservative, geometric-bridge, and hybrid prompt strategies, producing 10 unique recipes. \code{prompt\_00} (batch 32, dropout 0.4, learning rate 0.05, momentum 0.75) was selected by its three-seed one-epoch validation mean of 0.5247 and then evaluated for five-epoch persistence. No corresponding stronger-boundary contrastive result was reported, which explains the dash in Table S9. This route is exploratory and does not support a headline claim. The five-epoch means were:

\begin{table}[ht]
\centering
\begin{tabular}{lrrrrr}
\toprule
Target & Control & Copy & Existing target-only & Sobol & Contrastive \\
\midrule
Weak CIFAR-10 target & 0.4440 & 0.7411 & 0.7731 & 0.7531 & 0.7465 \\
Stronger CIFAR-10 boundary & 0.6815 & 0.7650 & 0.7279 & 0.7447 & -- \\
\bottomrule
\end{tabular}
\suppcaption{Five-epoch validation means from the completed overnight exploration. The weak-target and stronger-boundary labels refer to the CIFAR-10 target and architecture-related source records listed in Section 2. This is a 16-recipe validation-only search, not a completed 64-trial Optuna comparison.}
\end{table}

\subsection{Source-quality stress test}

The source-quality v5 analysis uses one weak CIFAR-10 target and four same-family source levels. It contains 15 arms, 120 attempts, 117 evaluations, and three failures, with zero official-test access. The source accuracy levels are 0.2423, 0.4311, 0.6718, and 0.8129. Mean arm-best accuracies are 0.4914, 0.3232, 0.4925, and 0.4903, respectively, compared with the control mean arm-best of 0.4949. The best-arm deltas are approximately $-0.0035$, $-0.1717$, $-0.0023$, and $-0.0046$. No reliable monotonic source-quality benefit is observed; 22 of 24 raw outputs and final recipes were identical across source levels, and prompt/transform choices remain confounded.

\subsection{Matched held-out replay}

A predeclared, replay-verified held-out analysis used 15 official-test accesses without new training. Weighted means were 0.1456 for the target control, 0.3303 for direct copy, 0.4615 for the existing target-only LLM recipe, and 0.4053 for the source-guided nominal draw. The existing target-only LLM recipe beat direct copy on every seed, and source-guided beat direct copy on every seed, but source conditioning did not exceed the existing target-only recipe in this matched setting. This is a negative control for universal source-conditioning claims, not a new headline comparison.

\section{Artifact provenance}

The source archive contains the ECCV main source, bibliography, style files, figure assets, and this supplement. The artifact archive contains the exact headline prompt JSON files, the 36 retrieved portfolio-detail records, \code{main\_candidate\_attempts.csv}, \code{matched\_count\_resampling.csv}, \code{candidate\_distribution\_data.csv}, the distribution figure, the deterministic analysis script, and SHA-256 manifests, together with the earlier non-sensitive evidence files used for the camera-ready analysis. The artifact README documents the workstation source paths and both CSV schemas. Protocol definitions and exact LEMUR record identifiers are provided in this supplement; raw launch manifests and the cleaned implementation are not part of the release. All numeric claims in the paper are either directly present in these files or are arithmetic differences stated in the corresponding tables.